# Solving Linear Equations Using a Jacobi Based Time-Variant Adaptive Hybrid Evolutionary Algorithm

**A.R. M. Jalal Uddin Jamali**
Department of Mathematics
Khulna University of Engineering & Technology (KUET)
Khulna-9203, Bangladesh
E-mail: jamali@math.kuet.ac.bd

**M. M. A. Hashem**
Department of Computer Science and Engineering
Khulna University of Engineering & Technology (KUET)
Khulna-9203, Bangladesh
E-mail:  hashem@cse.kuet.ac.bd

**Md. Bazlar Rahman**
Department of Mathematics
Khulna University of Engineering & Technology (KUET)
Khulna-9203, Bangladesh
Ph: +880-41-769471 Ext-523, 524

**Abstract**
*Large set of linear equations, especially for sparse and structured coefficient (matrix) equations, solutions using classical methods become arduous. And evolutionary algorithms have mostly been used to solve various optimization and learning problems. Recently, hybridization of classical methods (Jacobi method and Gauss-Seidel method) with evolutionary computation techniques have successfully been applied in linear equation solving. In the both above hybrid evolutionary methods, uniform adaptation (UA) techniques are used to adapt relaxation factor. In this paper, a new Jacobi Based Time-Variant Adaptive (JBTVA) hybrid evolutionary algorithm is proposed. In this algorithm, a Time-Variant Adaptive (TVA) technique of relaxation factor is introduced aiming at both improving the fine local tuning and reducing the disadvantage of uniform adaptation of relaxation factors. This algorithm integrates the Jacobi based SR method with time variant adaptive evolutionary algorithm. The convergence theorems of the proposed algorithm are proved theoretically. And the performance of the proposed algorithm is compared with JBUA hybrid evolutionary algorithm and classical methods in the experimental domain. The proposed algorithm outperforms both the JBUA hybrid algorithm and classical methods in terms of convergence speed and effectiveness.*

**KeyWords**
Adaptive algorithm, Evolutionary algorithm, Time-variant adaptation, Linear equations, Successive Relaxation, Mutation.

## INTRODUCTION

Solving a set of simultaneous linear equations is a fundamental problem that occurs in diverse applications. Linear system of equations are associated with many problems in engineering and science, as well as with applications of mathematics to the social sciences and the quantitative study of business, statistics and economic problems. Even the most complicated situations are frequently approximated by a linear model as a first step. Further, the solution of system of nonlinear equations is achieved by an iterative procedure involving the solution of a series of linear equations, each of them approximating the nonlinear equations. Similarly, the solution of ordinary differential equations, partial differential equations and integral equations using finite difference method lead to system of linear or nonlinear equations. Linear equations also arise frequently in numerical analysis [1,2]. After invent of easily accessible computers, one of the main issue how can increase the speed to solve equations. Also it is sometime desired to get a rapid solution of the physical problems for appropriate decision. For example, short-term weather forecast, image processing, simulation to predict aerodynamics performance which of these applications involve the solution of very large sets of simultaneous equations by numerical methods and time is an important factor for practical application of the results [3]. If the algorithm of solving equations can be implemented efficiently in parallel processing environment, it can easily decrease a significance time to get the result.

They are so many classical numerical methods to solve linear equations. For large number of linear equations, especially for sparse and structured coefficient (matrices) equations, iterative methods are preferable as iterative method are unaffected by round off errors to a large extent [4]. The well-known classical numerical iterative methods are the Jacobi method and Gauss-Seidel method. The rate of convergence, as very slow for both cases, can be accelerated by using SR technique [1,2]. But the speed of convergence depends on relaxation factor $\omega$ with a necessary condition for the convergence is $0 < \omega < 2$ [3,4]. However, it is often very difficult to estimate the optimal relaxation factor, which is a key parameter of the SR technique [1,5]. Anyway the Evolutionary Algorithms (EA) are stochastic algorithms whose search methods model some natural phenomena: genetic inheritance and Darwinian strife for survival [6,7,8]. Almost all of the works on EA can be classified as evolutionary optimization (either numerical or combinatorial) or evolutionary learning. But Fogel and Atmar [9] used linear equation solving as test problems for comparing recombination, inversion operations and Gaussian mutation in an evolutionary algorithm. However, they emphasized their study not on equation solving, but rather on comparing the effectiveness of recombination relative to mutation. No comparison with classical equation-solving methods was given. Recently, hybrid evolutionary algorithms [10,11] are developed by integrating classical SR technique based on Gauss-Seidel method and based on Jacobi method to solve equations in which the relaxation factor, $\omega$, is self-adapted by using uniform adaptation technique. Also obvious biological evidence is that a rapid change is observed at early stages of life and a slow change is observed at latter stages of life in all kinds of animals/plants. These changes are more often occurred dynamically depending on the situation exposed to them. By mimicking this emergent natural evidence, a special dynamic Time-Variant Mutation (TVM) operator is proposed by Hashem [12] and Michalewicz et al. [13,14, 15] in global optimization problems.

In this paper, a new hybrid algorithm is proposed in which time variant adaptive evolutionary computation techniques and SR technique are used in classical Jacobi method. The proposed Jacobi-based Time Variant Adaptive (JBTVA) hybrid algorithm does not require a user to guess or estimate the optimal relaxation factor $\omega$. The proposed algorithm initializes uniform relaxation factors in a given domain and "evolves" it by time variant adaptation technique instead of uniform adaptation technique. The proposed algorithm integrates the Jacobi-based SR method with evolutionary computation techniques, which uses initialization, recombination, mutation, adaptation, and selection mechanisms. It makes better use of a population by employing different equation-solving strategies for different individuals in the population. Then these individuals can exchange information through recombination and the error is minimized by mutation and selection mechanisms. Experimental results show that the proposed Jacobi-based time variant adaptive hybrid algorithm can solve linear equations quickly and efficiently compared to both the classical methods and the Jacobi based uniform adaptive hybrid method. Also this proposed algorithm can be implemented inherently in parallel processing environment efficiently.

## THE BASIC EQUATION OF JACOBI BASED SR METHODS

A linear system can be expressed as a matrix equation in which each matrix or vector element belongs to a field, typically the real number $\Re$. A set of linear equations in $n$ unknowns $x_1, x_2, \cdots, x_n$ is given by the matrix-vector equations:

$$\begin{bmatrix} a_{11} & a_{12} & \cdots & a_{1n} \\ a_{21} & a_{22} & \cdots & a_{2n} \\ \vdots & \vdots & \ddots & \vdots \\ a_{n1} & a_{n2} & \cdots & a_{nn} \end{bmatrix} \begin{bmatrix} x_1 \\ x_2 \\ \vdots \\ x_n \end{bmatrix} = \begin{bmatrix} b_1 \\ b_2 \\ \vdots \\ b_n \end{bmatrix}$$

(1)

or, equivalently, letting matrix $\mathbf{A} = (a_{ij}) \in \Re^n \times \Re^n$ and vector $\mathbf{x} = (x_i) \in \Re^n$, $\mathbf{b} = (b_i) \in \Re^n$ where $\Re$ is real number, as

$$\mathbf{Ax} = \mathbf{b}$$

(2)

For the solution of the linear Eq. (1), Jacobi method by using SR technique [1,2] in Eq. (1) is given by

$$x_i^{(k+1)} = x_i^{(k)} + \frac{\omega}{a_{ii}} \left( b_i - \sum_{j=1}^{n} a_{ij} x_j^{(k+1)} \right),$$

$i = 1, 2, \cdots, n$  and  $k = 0, 1, \cdots$

(3)

Now coefficient matrix $\mathbf{A}$ of the Eq. (2) can be decomposed as

$$\mathbf{A} = \mathbf{D} - \mathbf{L} - \mathbf{U}$$

(4)

where $\mathbf{D} = (d_{ij})$ is a diagonal matrix, $\mathbf{L} = (l_{ij})$ is a lower strictly triangular matrix and $\mathbf{U} = (u_{ij})$ is a upper strictly triangular matrix.

So in matrix form Eq.(3) can be rewrite as:

$$\mathbf{x}^{(k+1)} = \mathbf{H}_\omega \mathbf{x}^{(k)} + \mathbf{V}_\omega$$

(5)

where $\mathbf{H}_\omega = \mathbf{D}^{-1} \{\omega \mathbf{L} + (1-\omega)\mathbf{I} - \omega \mathbf{U}\}$ and $\mathbf{V}_\omega = \omega \mathbf{D}^{-1} \mathbf{b}$;

(6)

Here $\mathbf{H}_\omega$ is called Jacobi iteration matrix, $\mathbf{I}$ is called identity matrix and $\omega \in (\omega_L, \omega_U)$ is called relaxation factor which influence the convergence rate of the both methods greatly; $\omega_L$ and $\omega_U$ are denoted as lower and upper boundary values of $\omega$. The optimal relaxation factor has been discussed for some special matrix [1,5]. But, in general, it is very difficult to estimate the prior optimal relaxation factor.

## THE PROPOSED HYBRID ALGORITHM

The key idea behind the hybrid algorithm, that combines the Jacobi-based SR method with time variant adaptive evolutionary computation techniques, is to self-adapt the relaxation factor used in SR technique. For different individuals in a population, different relaxation factors are used to solve equations. The relaxation factors will be adapted based on the fitness of individuals (i.e. based on how well an individual solves the equations). Similar to many other evolutionary algorithms, the proposed hybrid algorithm always maintains a population of approximate solution to linear equations. Each solution is represented by an individual. The initial population is generated randomly form the field $\Re^n$. Different individuals use different relaxation factors. Recombination in the hybrid algorithm involves all individuals in a population. If the population size is $N$, then the recombination will have $N$ parents and generates $N$ offspring through linear combination. Mutation is achieved by performing one iteration of Jacobi method using SR technique as given by Eq. (5). The mutation is stochastic since $\omega$ used in the iteration is initially generated between $\omega_L$ (=0) and $\omega_U$ (=2) and $\omega$ is adapted stochastically in each generation (iteration) and adaptation nature of $\omega$ is also time variant. The fitness of an individual is evaluated based on the error of an approximate solution. For example, given an approximate solution (i.e., an individual) **z**, its error is defined by $\|e(\mathbf{z})\| = \|\mathbf{Az} - \mathbf{b}\|$. The relaxation factors are adapted after each generation, depending on how well an individual performs (in term of error). The main steps of the Jacobi-based hybrid evolutionary algorithm described as follows [10,11]:

**Step 1: Initialization**

Generate, randomly from $\Re^n$, an initial population of approximate solutions to the linear Eq.(1) using different relaxation factor for each individual of the population. Denote the initial population as $\mathbf{X}^{(0)} = \{\mathbf{x}_1^{(0)}, \mathbf{x}_2^{(0)}, \ldots, \mathbf{x}_N^{(0)}\}$ where $N$ is the population size. Let $k \leftarrow 0$ where $k$ is the generation counter. And initialize corresponding relaxation factor $\omega$ as:

$$\omega_i = \begin{cases} \omega_L + \dfrac{d}{2} & \text{for } i = 1 \\ \omega_{i-1} + d & \text{for } 1 < i \leq N \end{cases},$$

where $d = \dfrac{\omega_U - \omega_L}{N}$

(7)

**Step 2: Recombination**

Now generate $\mathbf{X}^{(k+c)} = \{\mathbf{x}_1^{(k+c)}, \mathbf{x}_2^{(k+c)}, \ldots, \mathbf{x}_N^{(k+c)}\}$ as an intermediate population through the following recombination:

$$\mathbf{X}^{(k+c)} = \mathbf{R}(\mathbf{X}^{(k)})^t$$

(8)

Where $\mathbf{R} = (r_{ij})_{N \times N}$ is a stochastic matrix [16], and the superscript $t$ denotes transpose.

**Step 3: Mutation**

Then generate the next intermediate population $\mathbf{X}^{(k+m)}$ from $\mathbf{X}^{(k+c)}$ as follows: For each individual $\mathbf{x}_i^{(k+c)}$ ($1 \leq i \leq N$) in population $\mathbf{X}^{(k+c)}$ produces an offspring according to Eq. (5)

$$\mathbf{x}_i^{(k+m)} = \mathbf{H}_{\omega_i} \mathbf{x}_i^{(k+c)} + \mathbf{V}_{\omega_i}, \quad i = 1, 2, \ldots, N.$$

(9)

Where $\omega_i$ is denoted as relaxation factor of the *i*th individual and $\mathbf{x}_i^{(k+m)}$ is denoted as *i*th (mutated) offspring, so that only one iteration is carried out for each mutation.

**Step 4: Adaptation**

Let $\mathbf{x}^{(k+m)}$ and $\mathbf{y}^{(k+m)}$ be two offspring individuals with relaxation factors $\omega_x$ and $\omega_y$ and with errors (fitness value) $\|e(\mathbf{x}^m)\|$ and $\|e(\mathbf{y}^m)\|$ respectively. Then the relaxation factors $\omega_x$ and $\omega_y$ are adapted as follows:

(a) If $\|e(\mathbf{x}^m)\| > \|e(\mathbf{y}^m)\|$, (i) then move $\omega_x$ toward $\omega_y$ by using

$$\omega_x^m = (0.5 + p_x)(\omega_x + \omega_y)$$

(10)

and (ii) move $\omega_y$ away from $\omega_x$ using

$$\omega_y^m = \begin{cases} \omega_y + p_y(\omega_U - \omega_y), & \text{when } \omega_y > \omega_x \\ \omega_y + p_y(\omega_L - \omega_y), & \text{when } \omega_y < \omega_x \end{cases}$$

(11)

Where $p_x = E_x \times N(0, 0.25) \times T_\omega$, $p_y = E_y \times N(0, 0.25) \times T_\omega$, and are denoted as Time-Variant Adaptive (TVA) probability parameter of $\omega_x$ and $\omega_y$ respectively.

Here $T_\omega = \lambda \ln(1 + \dfrac{1}{t + \lambda})$, $\lambda > 10$

(12)

Which is the Basic Time-Variant (BTV) parameter in which $\lambda$ is an exogenous parameter, used for increased or decreased of rate of change of curvature with respect to number of iterations, *t*. Also $N(0, 0.25)$ is the Gaussian distribution with mean 0 and standard deviation 0.25. Now $E_x$ and $E_y$ denote the approximate initial boundary of the variation of TVA parameters of $\omega_x$ i.e. $(-E_x, E_x)$ and $\omega_y$ i.e. $(-E_y, E_y)$ respectively. And $\omega_x^m$ & $\omega_y^m$ are adapted relaxation factors correspond to $\omega_x$ and $\omega_y$.

(b) If $\|e(\mathbf{x}^m)\| < \|e(\mathbf{y}^m)\|$, then adapt $\omega_x$ and $\omega_y$ in the same way as above but reverse the order of $\omega_x^m$ and $\omega_y^m$.

(c) If $\|e(\mathbf{x}^m)\| = \|e(\mathbf{y}^m)\|$, no adaptation. So that $\omega_x^m = \omega_x$ and $\omega_y^m = \omega_y$.

**Step 5: Selection and Reproduction**

Select the best N/2 offspring individuals according to their fitness values (errors). Then reproduce of the above selected offspring (i.e. each parents individual generates two offspring). Then form the next generation of N individuals.

**Step 6: Termination**

If $min\{\|e(\mathbf{z})\| : \mathbf{z} \in \mathbf{X}\} < \eta$ (Threshold error), then stop the algorithm and get unique solution. If $min\{\|e(\mathbf{z})\| : \mathbf{z} \in \mathbf{X}\} \to \infty$, then stop the algorithm but fail to get any solution. Otherwise go to Step 2.

**THEOREMS**

The following theorem establishes the rapid convergence of the hybrid algorithms.

*Theorem-1*: If there exist an $\varepsilon (0 < \varepsilon < 1)$ such that, for the norm of $\mathbf{H}_\omega$,

$$\|\mathbf{H}_\omega\| < \varepsilon < 1, \text{ then } \lim_{k \to \infty} \mathbf{x}^{(k)} = \mathbf{x}^*,$$

*where $\mathbf{x}^*$ is the solution vector to the system of linear equations.*

**Proof**: The proof of this theorem simple and straightforward and proof of this theorem is given in [10, 11].

The following theorem justifies the adaptation technique for relaxation factors used in proposed hybrid evolutionary algorithms.

***Theorem –2:*** Let $\rho(\omega)$ be the spectral radius of matrix $\mathbf{H}_\omega$, $\omega^*$ be the optimal relaxation factor, and let $\omega_x$ and $\omega_y$ are the relaxation factors of the selected pair individuals $x$ and $y$ respectively. Assume $\rho(\omega)$ is monotonic decreasing when $\omega < \omega^*$ and $\rho(\omega)$ is monotonic increasing when $\omega > \omega^*$. Also consider $\rho(\omega_x) > \rho(\omega_y)$. Then

(i) $\rho(\omega_x^m) < \rho(\omega_x)$, when
$$\omega_x^m = (0.5 + p_x)(\omega_x + \omega_y)$$ where $p_x \in [-E_x, E_x]$ and

(ii) There is a very high probability that $\rho(\omega_y^m)$ is less than $\rho(\omega_y)$

i.e $\rho(\omega_y^m) < \rho(\omega_y)$ when

$$\omega_y^m = \begin{cases} \omega_y + p_y(\omega_U - \omega_y), & \text{when } \omega_y > \omega_x \\ \omega_y + p_y(\omega_L - \omega_y), & \text{when } \omega_y < \omega_x \end{cases}, \text{ where}$$

$p_y \in [0, E_x]$.

**Proof**: The first result can be derived directly from the monotonicity of $\rho(\omega)$. The second result can also be derived from the monotonicity of $\rho(\omega)$ with a very high probability as [11].

## PERFORMANCE OF THE HYBRID ALGORITHM

In order to evaluate the effectiveness of the proposed JBTVA hybrid algorithm, numerical experiments have been carried out on a number of problems to solve the systems of linear Eq. (1) of the form: $\mathbf{Ax} = \mathbf{b}$

The following settings are valid all through the experiments:

The dimension of unknown variable is $n = 100$, population size $N = 2$, boundary of relaxation factors ($\omega_L$, $\omega_U$) = (0,2), (i.e. only two individuals were used so that initial $\omega$'s become 0.5 and 1.5 respectively) the approximate initial boundary, $E_x$ and $E_y$ are set at 0.125 and 0.03125 respectively the exogenous parameter $\lambda$ is set at 50, each individual $\mathbf{x}$ of population $\mathbf{X}$ is initialized from the domain $\Re^{100} \in (-30, 30)$ randomly and uniformly and the stochastic matrix $\mathbf{R}$ was chosen as follows:

If the fitness of the first individuals was better then the second, let

$$\begin{pmatrix} \mathbf{x}_1^{(k+c)} \\ \mathbf{x}_2^{(k+c)} \end{pmatrix} = \begin{pmatrix} 1.0 & 0 \\ 0.99 & 0.01 \end{pmatrix} \begin{pmatrix} \mathbf{x}_1^{(k)} \\ \mathbf{x}_2^{(k)} \end{pmatrix}$$

(13)

else let

$$\begin{pmatrix} \mathbf{x}_1^{(k+c)} \\ \mathbf{x}_2^{(k+c)} \end{pmatrix} = \begin{pmatrix} 0.01 & 0.99 \\ 1.0 & 0.0 \end{pmatrix} \begin{pmatrix} \mathbf{x}_1^{(k)} \\ \mathbf{x}_2^{(k)} \end{pmatrix}$$

14)

Each experiment is run 10 times using 10 different sample paths and then averaged them.

Now the first problem is to solve linear equations, Eq. (1), where $a_{ii}$ = (-70,70); $a_{ij}$ = (0,7); $b_i$ = (0,70), $i, j = 1, \cdots, n$ (i.e. problem $P_2$ in Table I). A single set of parameters are generated randomly from the above mentioned problem and the following experiments are carried out. The problem is to be solved with an error smaller than $10^{-6}$ (threshold error). **Fig.** 1 shows the numerical results (in graphical form) achieved by the proposed classical Jacobi based SR method with several relaxation factors ($\omega$ = 0.05, 0.5, 1.0 and 1.5) and proposed JBTVA hybrid algorithm. And **Fig**. 2 shows the numerical results (in graphical form) achieved by the proposed classical Gauss-Seidel based SR method with several relaxation factors ($\omega$ = 0.05, 0.5, 1.0 and 1.5) and proposed hybrid algorithm.

It is observed in Figure 1 and Figure 2 that the rate of convergence of JBTVA algorithm is better than that of both classical Jacobi based SR method and Gauss-Seidel based SR method except for $\omega$ = 0.5 where Gauss-Seidel based SR method converges a bit fast than JBTVA method. It is also observed that both classical methods are sensitive to the relaxation factors whereas JBTVA algorithm is not so.

Figure 3 also shows the numerical results (in graphical form) achieved by the proposed hybrid algorithm and JBUA (Jacobi based Uniform Adaptation) [11] hybrid algorithm. It is observed in Figure 3 that the rate of convergence of TVA-based algorithm is better than that of UA-based algorithm.

**Table** I presents ten test problems, labeled from $P_1$ to $P_{10}$, with dimension, $n = 100$. For each test problem $P_i$: $i$ = 1, 2, . . ., 10, the coefficient matrix $\mathbf{A}$ and constant vector $\mathbf{b}$ are all generated uniformly and randomly within given domains (shown in 2nd column with corresponding rows of Table I. This table shows the comparison of the number of generation (iteration) of the JBUA and proposed JBTVA hybrid algorithms to the given preciseness, $\eta$ (see column three of the Table I). One observation can be made immediately from this table, except for problem $P_{10}$ where the JBUA algorithm performed near to same as JBTVA algorithm, TVA-based hybrid algorithm performed much better than the UA-based hybrid algorithm for all other problems.

Figure 4 shows the nature of self-adaptation of $\omega$ in the UA-based hybrid algorithm and Figure 5 shows the nature of self-adaptation of $\omega$ in the TVA-based hybrid algorithm. It is observed in Figure 4 and Figure 5 that the self-adaptation process of relaxation factors in TVA-based hybrid algorithm is much better than that of in UA-based hybrid algorithm. Fig. 5 shows that how initial $\omega = 0.5$, is adapted to its near optimum value and reaches to a better position for which rate of convergence is accelerated. On the other hand Figure 4 shows that initially $\omega = 0.5$, by self-adaptation process, does not gradually reaches to a better position.

## PARALLEL PROCESSING

The parallel searching is one of the main properties of evolutionary computational techniques. Now since classical Jacobi based SR method can be implemented in parallel processing environment [2,3]. So JBTVA, as like as JBUA [11], can also be implemented, inherently, in parallel processing environment efficiently. Where as Gauss-Seidel based hybrid algorithm [10], inherently, can not be implemented in parallel processing environment efficiently.

## CONCLUDING REMARKS

In this paper, a Time-variant adaptive (TVA)-based hybrid evolutionary algorithm has been proposed for solving systems of linear equations. The TVA-based hybrid algorithm integrates the classical Jacobi based SR method with evolutionary computation techniques. The time-variant based adaptation is introduced for adaptation of relaxation factors, which makes the algorithm more natural and accelerates its rate of convergence. The recombination operator in the algorithm mixed two parents by a kind of averaging, which is similar to the intermediate recombination often used in evolution strategies [6,7]. The mutation operator is equivalent to one iteration in the Jacobi based SR method. The mutation is stochastic and time variant since the relaxation factor $\omega$ is adapted stochastically. The proposed TVA-based relaxation factor $\omega$ adaptation technique acts as a local fine tuner and helps to escape from the disadvantage of uniform adaptation. The effectiveness of this hybrid algorithm is compared with that of classical Jacobi based SR method and Gauss-Seidel based SR method. Also numerical experiments with various test problems have shown that the proposed JBTVA hybrid algorithm performs better than the JBUA hybrid algorithm. This preliminary investigation has showed that this algorithm outperforms JBUA hybrid algorithm as well as classical SR methods. Jacobi-based hybrid algorithm is also very simple and easy to implement both in sequential and parallel computing environment.